# Research on Autonomous Driving Decision-making Strategies based Deep Reinforcement Learning

Zixiang Wang *

College of Engineering and Computer Science, Syracuse University, Syracuse, NY, USA, zwang161@syr.edu

Hao Yan

College of Engineering and Computer Science, Syracuse University, Syracuse, NY, USA, hyan17@syr.edu

Changsong Wei

College of Engineering and Computer Science, Independent Researcher, Syracuse, Chengdu, Sichuan, China, changsongwei88@gmail.com

Junyu Wang

UCLA Computer Science Department, University of California, Los Angeles, California, MA, USA, junyu08@cs.ucla.edu

Minheng Xiao

Department of Integrated System Engineering, Ohio State University, Columbus, OH, USA, minhengxiao@gmail.com

**Abstract**

The behavior decision-making subsystem is a key component of the autonomous driving system, which reflects the decision-making ability of the vehicle and the driver, and is an important symbol of the high-level intelligence of the vehicle. However, the existing rule-based decision-making schemes are limited by the prior knowledge of designers, and it is difficult to cope with complex and changeable traffic scenarios. In this work, an advanced deep reinforcement learning model is adopted, which can autonomously learn and optimize driving strategies in a complex and changeable traffic environment by modeling the driving decision-making process as a reinforcement learning problem. Specifically, we used Deep Q-Network (DQN) and Proximal Policy Optimization (PPO) for comparative experiments. DQN guides the agent to choose the best action by approximating the state-action value function, while PPO improves the decision-making quality by optimizing the policy function. We also introduce improvements in the design of the reward function to promote the robustness and adaptability of the model in real-world driving situations. Experimental results show that the decision-making strategy based on deep reinforcement learning has better performance than the traditional rule-based method in a variety of driving tasks.

## 1 INTRODUCTION

With the rapid development of intelligent network technology, the process of global automotive intelligence is accelerating. Intelligent networked vehicles are becoming an important direction for the development of the domestic and foreign automobile industry by virtue of their significant advantages in reducing safety accidents, reducing environmental pollution, alleviating traffic congestion and reducing energy consumption. Self-driving cars follow a predetermined route, but the real traffic environment is not static. Roads are often filled with a variety of traffic participants, whose real-time status is constantly changing. Therefore, autonomous vehicles need to dynamically predict the surrounding environment and implement motion interaction strategies such as following, avoiding, and changing lanes according to different traffic participants [1]. Not only does this help to adjust the driving strategy in real time, but it also provides guidance for trajectory planning, ensuring that the vehicle can complete the driving task as efficiently and safely as possible.

The autonomous driving system can be regarded as a combination of traditional automotive technology and modern high-performance computer technology, covering environmental perception, map positioning, prediction, decision-making, planning and control execution. At the perception and positioning layer, autonomous vehicles are equipped with a variety of sensors such as lidar, millimeter-wave radar, and cameras, and through sensor fusion technology combined with high-precision maps, they can comprehensively perceive the static and dynamic environment around the vehicle, and realize high-precision positioning and global path planning [2].

At the level of decision-making and planning, the system includes three aspects: the trajectory prediction of traffic participants, the behavior decision-making of the vehicle and the trajectory planning, so as to generate a driving trajectory that is both safe and collision-free and satisfies specific constraints. Finally, in the control layer, the system uses the vehicle dynamics control theory to output control instructions such as accelerator, brake, and steering wheel angle, and realizes accurate control of the vehicle movement through the wire-controlled chassis to ensure that the actual driving trajectory of the vehicle can accurately track the planned trajectory [3].

At present, behavioral decision-making methods are mainly divided into rule-based methods and data-driven learning methods. Rule-based decision-making scenarios rely on pre-established traffic rules and driving experience. However, this approach cannot exhaust all possible scenarios because real-world traffic scenarios are often extremely complex, and dynamic traffic participants may act in exactly opposite directions in an instant [4]. Therefore, when faced with unknown scenarios, rule-based decision-making systems may encounter processing power limitations, and may require driver intervention to deal with complex situations.

With the development of artificial intelligence technology, especially the success achieved in areas such as image recognition, financial market analysis, and natural language processing, more solutions are provided for autonomous driving technology. Among them, reinforcement learning methods can enable agents to make more reasonable behavioral decisions according to environmental changes through automatic interaction and continuous optimization with the environment [5]. Combined with the powerful representation ability of deep learning neural networks, reinforcement learning provides an important way for the realization of autonomous driving decision-making technology, especially when dealing with high-dimensional complex problems, which can significantly improve the intelligence level of decision-making systems.

The behavioral decision-making subsystem is a core part of the complex and dynamic urban traffic environment, and its main tasks are to ensure the safety of autonomous vehicles, comply with traffic laws, and provide the constraint information needed for smooth route and speed optimization. Fast and accurate behavioral decision-making can effectively reduce the occurrence of accidents, which is of great significance to improve personal safety. Having decision-making capabilities similar to those of drivers is a key sign of high-level intelligence in automobiles, and it also directly determines the level of autonomous driving systems [6].

Depending on the architecture, the input information of the decision-making system may include structured road information including the current lane, merging lane, intersection, and etc. Traffic sign information and obstacle status and trajectory prediction information of surrounding traffic participants. Through a comprehensive analysis of the current environment and traffic participant information, the system can output driving action instructions including following the car, changing lanes freely, forcing lane changing, and overtaking [7].

## 2  RELATED WORK

In the field of autonomous driving decision-making technology, there are currently two main types of decision-making methods: rule-based methods and learning-based methods. The rule-based decision-making method is characterized by its strong logic, which relies on the driver's behavior rule library constructed by driving experience and traffic rules. These methods select specific driving behaviors through logical judgments about the surrounding environment. Typical rule models include finite state machines and decision trees. The finite state machine is composed of a plurality of driving states and the transition relationship between them, which will lead to the transfer of states when the vehicle or environment changes, so as to generate the corresponding driving action. The decision tree model is a tree-like decision-making analysis tool, which constructs a tree-shaped diagram in the form of lines and graphs such as strategy, state, probability, and return value. The leaf nodes of the decision tree represent specific driving behavior strategies, while each non-leaf node represents the judgment of the driving state. Therefore, a decision tree can be transformed into a complex "If-Then" logical judgment rule. When making decisions, the search for policies is typically conducted through a top-down "polling" mechanism [8].

The rule-based behavior decision-making model has the advantages of strong logic and easy editing, so it has been widely used in the current behavior decision-making subsystem of autonomous vehicles. However, when the scene environment becomes complex, the limitations of the rule-based approach become apparent. The factors to be considered in decision rules have increased dramatically, resulting in a more complex decision tree structure and more difficult to extract and divide the state transition conditions of finite state machines. These challenges make rule-based decision-making models difficult to cope with dynamic and complex traffic scenarios. Rule-based decision-making models usually include a follow-up model and a lane change model, which are used to deal with horizontal and vertical decision-making judgments. The traditional following model is generally based on the theory of vehicle kinematics and derives a model with definite physical significance. These models are designed to simulate the behavior characteristics and response patterns of a vehicle when following the vehicle in front.

In the study of following and changing lanes, Gazis et al. [9] conducted experiments on the relationship between the following vehicle in front and the response of the self-vehicle by analyzing the traffic flow and density. In their study, they introduced the relationship between vehicle pitch and speed change, and established a stimulus-response Gazis-Herman-Rothery (GHR) nonlinear following model. The model reveals how changes in the motion of the vehicle in front create a stimulus response to the vehicle behind, causing the vehicle behind to adjust its driving behavior. The GHR model not only considers the influence of the motion of the vehicle in front on the vehicle behind, but also describes the behavior of the vehicle in the complex traffic environment through a nonlinear relationship, which provides a more accurate theoretical basis and practical guidance for the following model.

The direct learning-based decision model can be seen as an evolved version of the rule-based model. This kind of model learns from a large number of human driving data to form a set of driving rule libraries based on "scene features-driving actions". Unlike traditional rule-based inference models, learning-based decision-making models not only rely on pre-set traffic rules and driving experience, but also continuously optimize and update the rule base through a data-driven approach. In practice, the system matches the current environment with the characteristics of the scene in the rule base, so as to select the most suitable driving action. This method is not only able to adapt to changing driving environments, but also has a certain degree of adaptability and optimization ability. By learning from real driving data, the model can generate decision-making strategies that are more in line with actual driving needs, and improve the rationality and safety of driving behavior. This learning-based decision-making method shows greater flexibility and adaptability than traditional rule models when dealing with complex and dynamic traffic scenarios.

In recent years, machine learning methods have been widely used in the field of artificial intelligence, and learning-based methods have gained more and more attention in academia and industry. For example, Mirchevska et al. [10] used the Deep Q-Network deep reinforcement learning method to deal with vehicle lane change decisions in highway scenarios. Compared with the traditional methods, simulation results show that the DQN method has obvious advantages in decision-making performance. In addition, the introduction of the SAC algorithm enables reinforcement learning agents to have stronger exploration capabilities and faster convergence effects.

## 3 METHODOLOGIES

### 3.1 Deep Q network

The DQN method is based on an approximation of the state-action-value function $Q(s, a)$, which evaluates the expected cumulative reward for taking action a in state s and continuing according to a specific strategy. DQN uses a deep neural network to approximate $Q(s, a)$. The Q-value function is updated by the Bellman equation, which is expressed as Equation 1.

$$Q(s, a) \leftarrow Q(s, a) + \alpha[r + \gamma max_{a'}Q(s', a') - Q(s, a)] \quad (1)$$

Where $\alpha$ is the learning rate and $r$ is the reward received after taking action $a$ in state $s$. $\gamma$ is the discount factor. $s'$ is the next state. $a'$ is an action in state $s'$. Neural network $Q(s, a; \theta)$ is parameterized by the parameter $\theta$, whose training goal is to minimize the loss function and expresses as Equation 2.

$$L(\theta) = \mathbb{E}_{s,a,r,s'}[(y - Q(s, a; \theta))^2] \quad (2)$$

Where $y = r + \gamma max_{a'}Q(s', a'; \theta^-)$ is the target value and $\theta^-$ is the parameter of the target network.

Deep Q Network (DQN) is a reinforcement learning method based on deep learning, which uses a deep neural network to approximate the state-action value function. The core idea of DQN is to guide agents to choose the optimal action strategy by learning the expected cumulative reward of state-action pairs. The method uses the Bellman equation to update the Q value, and trains the neural network by minimizing the difference between the target Q value and the current Q value. In the training process, DQN uses the experience replay mechanism to store and randomly sample past experience, combined with the target network to reduce the instability in training.

### 3.2 Proximal policy optimization

Above all, proposed proximal policy optimization improves the decision-making process by directly optimizing the policy function $\pi(a \mid s)$ rather than approximating the value function. PPO addresses the problem of policy updates by introducing a clipped objective function to ensure stable and reliable policy improvements. The PPO method limits the magnitude of each update when updating the policy, thus avoiding drastic changes in the policy, which makes the training process more stable and efficient. The objective function of PPO is expressed as Equation 3.

$$T^{PPO}(\theta) = \mathbb{E}_t [\min(\frac{\pi_\theta(a_t|s_t)}{\pi_{\theta old}(a_t|s_t)} \hat{A}_t, clip(\frac{\pi_\theta(a_t|s_t)}{\pi_{\theta old}(a_t|s_t)}, 1-\epsilon, 1+\epsilon)\hat{A}_t)] \qquad (3)$$

Where $\pi_\theta$ is the strategy defined by the parameter $\theta$. $\pi_{\theta old}$ is the previous strategy. $\hat{A}_t$ is the dominance function. $\epsilon$ is a clipping parameter.

The reason for introducing the advantage function $\hat{A}_t$ is that it can effectively reduce the variance and improve the efficiency of policy updates. The dominant function is calculated as follows Equation 4, where $\lambda$ is the Generalized Advantage Estimation (GAE) parameter, which controls the weight of the time difference. $\gamma$ is the discount factor. $V(s_t)$ is a function of the state value, which represents the expected return in the state $s_t$.

$$\hat{A}_t = \sum_{l=0}^{\infty} (\gamma\lambda)^l (r_{t+l} + \gamma V(s_{t+l}) - V(s_t)) \qquad (4)$$

The key to PPO lies in its objective function design, which effectively limits the magnitude of change between the old and new policies by introducing a shear term, $clip(\frac{\pi_\theta(a_t|s_t)}{\pi_{\theta old}(a_t|s_t)}, 1-\epsilon, 1+\epsilon)$, and avoids the instability caused by too fast policy updates. This design makes PPO more robust and efficient in strategy optimization.

Initially, a batch of data (state, action, reward, next state) is sampled from the environment, and then the dominance value $\hat{A}_t$ for each action is calculated using generalized advantage estimation (GAE). Next, the strategy parameter $\theta$ is updated by optimizing the objective function of PPO, and finally the state-value function $V(s)$ is updated by minimizing the value loss function. Through limiting the magnitude of policy updates, PPO ensures the stability and reliability of policy improvement, enabling agents to learn and optimize decision-making strategies more effectively in complex and dynamic environments.

### 3.3 Reward function

In order to improve the robustness and adaptability of the model, we have improved the design of the reward function. Designing a reasonable reward function is essential for reinforcement learning models to work effectively in complex traffic environments. We use a reward function in the form of Equation 5.

$$R(s,a) = w_1 \cdot R_{safety}(s,a) + w_2 \cdot R_{comfort}(s,a) + w_3 \cdot R_{efficiency}(s,a) \qquad (5)$$

Specifically, the reward function consists of the following three parts, each of which has its own specific role: safety rewards punish unsafe actions and encourage safe driving. The reward function in this part will punish the vehicle according to whether it violates traffic rules, whether it is close to a collision, etc. For example, when a vehicle approaches a vehicle in front or other obstacles, the safety reward is significantly reduced, guiding the agent to avoid risky behavior. Comfort Rewards Promote a smooth ride and improve passenger comfort. The reward function in this section takes into account factors such as changes in the vehicle's acceleration, hard braking, and sharp turns. By punishing these uncomfortable behaviors, the model is guided to choose smoother driving maneuvers, such as minimizing hard acceleration and hard braking. Efficiency Incentives Encourage energy-efficient driving and improve driving efficiency. This part of the reward function is rewarded based on factors such as fuel consumption, driving speed, and driving time. For example, vehicles are encouraged to drive at a more efficient speed under the premise of ensuring safety, avoiding unnecessary stopping and idling, so as to achieve the effect of energy conservation and emission reduction.

The weight $w_1$, $w_2$, $w_3$ is used to balance the rewards for the above three sections. The selection of these weights needs to be adjusted for the specific application to ensure that the model performs well in a wide range of driving conditions. Through the detailed design of the above steps, the reward function can guide the reinforcement learning model to learn safe, comfortable and efficient driving strategies in complex traffic environments. This not only improves the robustness and adaptability of the model, but also ensures the feasibility and reliability of the autonomous driving system in real-world applications.

During driving, you may encounter various unexpected events, such as pedestrians suddenly crossing the road, sudden braking of the vehicle in front, etc. These situations require the driving system to be able to react quickly and not just rely on a preset reward function. Different weather conditions have different effects on driving decisions. The system needs to be able to adjust driving strategies in different weather conditions to ensure safety and efficiency. Pavement conditions, such as potholes, standing water, construction, etc., can also affect driving decisions. The reward function needs to be able to dynamically adjust to different road conditions.

## 4 EXPERIMENTS

### 4.1 Experimental setups

In this experiments, we used the HighwayEnv environment for experiments. HighwayEnv is a highly configurable highway driving scenario simulator that provides a realistic test environment for testing a wide range of autonomous driving tasks. It supports a wide range of driving scenarios and provides a flexible API for easy integration with reinforcement learning frameworks. Following Figure 1 shows the used environment. For the DQN model, we use a three-layer convolutional neural network as the Q network, with the input layer being the state representation and the output layer being the action value. The learning rate is set at 0.001, the discount factor $\gamma$ is 0.99, the experience replay buffer size is 50000, and the batch size is 64. In addition, we update the target network every 1000 steps to ensure the stability of the training. For the PPO model, we used two fully connected feedforward neural networks as the strategy network and the value network, each containing two layers of 128 neurons in each layer. The learning rates of the policy network and the value network are set to 0.0003 and 0.001, respectively. The shear parameter $\varepsilon$ was set to 0.2 and the generalized advantage estimation parameter $\lambda$ was set to 0.95. With each update, we sample data for 2048 steps and optimize 10 times.

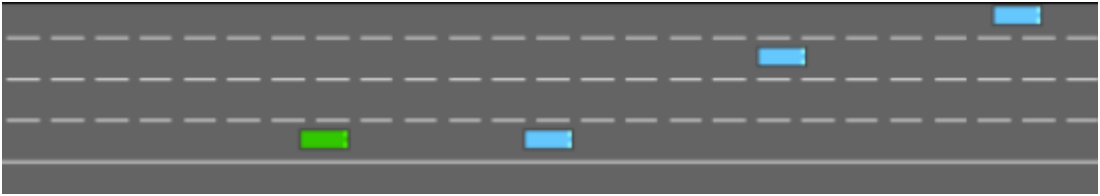

Figure 1. Illustration of Used HighwayEnv Environment.

### 4.2 Experimental analysis

In the HighwayEnv environment, the payback function balances the two goals of speed optimization and collision avoidance. Specifically, the reward function not only rewards the vehicle for maintaining an efficient and safe driving speed, but also imposes a strong penalty for collision avoidance, which guide the agent to ensure driving safety while pursuing speed. Figure 2 shows how the average return of the DQN algorithm in the highway merging task varies with the number of training rounds. The red curve represents the average return, and the shaded area indicates the standard deviation range of the return, reflecting the error of the return. As the number of training rounds increases, the average return gradually increases, indicating that the model is constantly learning and optimizing, and the performance is gradually improving. The design of this reward function balances speed optimization and collision avoidance, ensuring that the agent can drive safely while pursuing efficient driving.

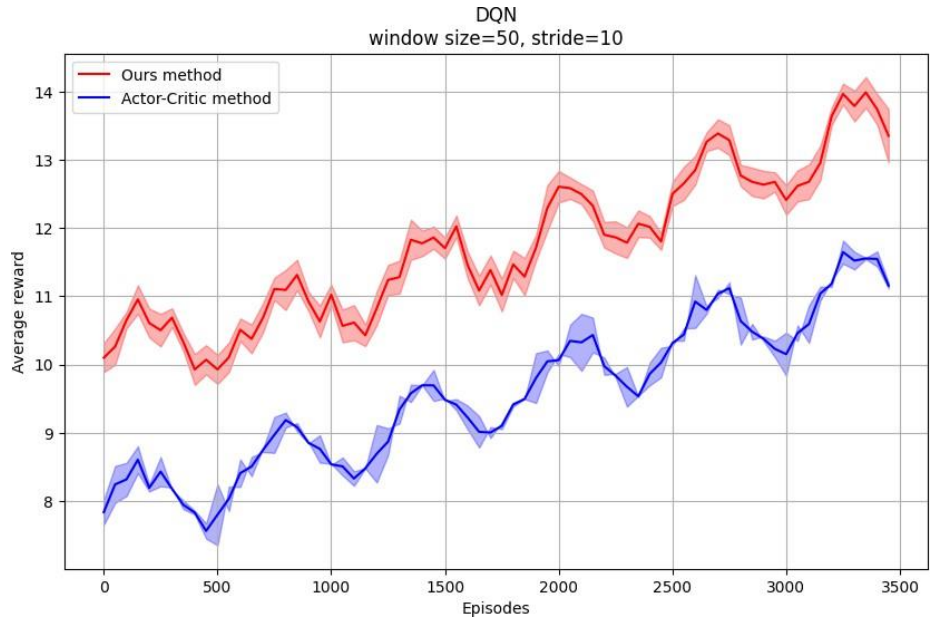

Figure 2. Average Return of DQN Algorithm Results.

Further, Figure 3 shows the path decision-making process of the green car. The yellow line indicates that the car enters from the left, changes lanes to the middle lane, then changes lanes to the right lane, and finally exits the screen. The path shows the behavior of the trolley in a highway merging and lane change scenario. Through such a path display, the driving decision-making process of the car can be intuitively observed and analyzed.

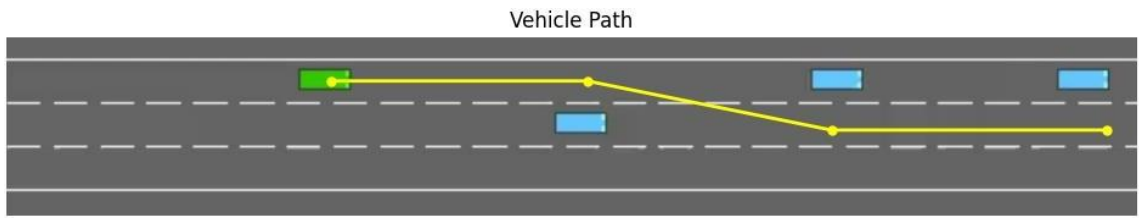

Figure 3. Demonstration of Path Decision-making Results.

The Figure 4 shows the experimental results of the system failure rate, including the number of failures and the duration of failures as a function of time step. The red curve represents the cumulative number of failures, and the blue curve represents the cumulative failure duration. From this graph, it can be observed that the number and duration of failures gradually increase over time. A lower failure rate and a shorter failure duration indicate a more stable system. By monitoring these indicators, the stability and reliability of the autonomous driving system can be effectively evaluated.

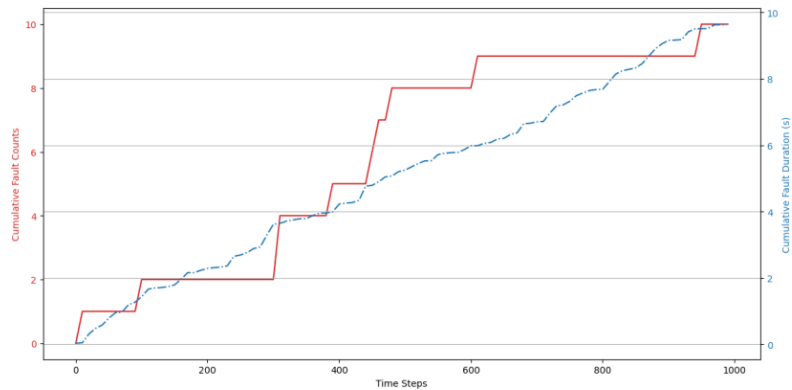

Figure 4. System Fault Rate: Fault Counts and Fault Duration Over Time.

## 5 CONCLUSION

In conclusion, by using Deep Q-Network (DQN) and Proximal Policy Optimization (PPO), autonomous vehicles can effectively learn and optimize driving strategies in complex traffic environments, surpassing the limitations of traditional rule-based methods. Comparative experiments show that DQN enables precise action selection, while PPO enhances decision-making quality. Improvements in the reward function design further increase the model's robustness and adaptability in real-world conditions. Overall, DRL-based strategies outperform traditional methods, highlighting their potential to revolutionize autonomous driving with higher intelligence and adaptability.

## REFERENCES

[1] Li, Guofa, et al. "Decision making of autonomous vehicles in lane change scenarios: Deep reinforcement learning approaches with risk awareness." Transportation research part C: emerging technologies 134 (2022): 103452.

[2] Li, Shen, et al. "Planning and decision-making for connected autonomous vehicles at road intersections: A review." Chinese Journal of Mechanical Engineering 34 (2021): 1-18.

[3] Omeiza, Daniel, et al. "Explanations in autonomous driving: A survey." IEEE Transactions on Intelligent Transportation Systems 23.8 (2021): 10142-10162.

[4] Badue, Claudine, et al. "Self-driving cars: A survey." Expert systems with applications 165 (2021): 113816.

[5] Huang, Zhiyu, Jingda Wu, and Chen Lv. "Efficient deep reinforcement learning with imitative expert priors for autonomous driving." IEEE Transactions on Neural Networks and Learning Systems 34.10 (2022): 7391-7403.

[6] Mariani, Stefano, Giacomo Cabri, and Franco Zambonelli. "Coordination of autonomous vehicles: Taxonomy and survey." ACM Computing Surveys (CSUR) 54.1 (2021): 1-33.

[7] Di, Xuan, and Rongye Shi. "A survey on autonomous vehicle control in the era of mixed-autonomy: From physics-based to AI-guided driving policy learning." Transportation research part C: emerging technologies 125 (2021): 103008.

[8] Zhu, Zeyu, and Huijing Zhao. "A survey of deep RL and IL for autonomous driving policy learning." IEEE Transactions on Intelligent Transportation Systems 23.9 (2021): 14043-14065.

[9] Gazis, Denos C., Robert Herman, and Richard W. Rothery. "Nonlinear follow-the-leader models of traffic flow." Operations research 9.4 (1961): 545-567.

[10] Mirchevska, Branka, et al. "High-level decision making for safe and reasonable autonomous lane changing using reinforcement learning." 2018 21st International Conference on Intelligent Transportation Systems (ITSC). IEEE, (2018): 2156-2162.